\documentclass{nle}

\usepackage{amsmath}
\usepackage{graphicx}
\usepackage{natbib}
\ifpdf%
\usepackage{epstopdf}%
\else%
\fi

\usepackage{url}
\usepackage{subfig}
\usepackage{multirow}
\usepackage{float}
\usepackage{graphicx}

\usepackage{caption}
\DeclareCaptionFont{captnfont}{\small\fontseries{n}\fontfamily{phv}\selectfont}
\begin{document}

\lefttitle{Meemi: Post-processing Cross-lingual Embeddings}
\righttitle{Natural Language Engineering}

\papertitle{Article}

\jnlPage{1}{00}
\jnlDoiYr{2020}
\doival{10.1017/x}

\title{Meemi: A Simple Method for Post-processing and Integrating Cross-lingual Word Embeddings}

\begin{authgrp}
\author{Yerai Doval}
\affiliation{Grupo COLE, Escola Superior de Enxe\~{n}ar\'ia Inform\'atica, \\Universidade de Vigo, Spain\\
        \email{yerai.doval@uvigo.es}}
\end{authgrp}

\begin{authgrp}
\author{Jose Camacho-Collados}
\author{Luis Espinosa-Anke}
\author{ Steven Schockaert}
\affiliation{School of Computer Science and Informatics,\\
       Cardiff University, UK\\
        \email{\{camachocolladosj,espinosa-ankel,schockaerts1\}@cardiff.ac.uk}}
\end{authgrp}



\begin{abstract}
Word embeddings have become a standard resource in the toolset of any Natural Language Processing practitioner.
While monolingual word embeddings encode information about words in the context of a particular language, cross-lingual embeddings define a multilingual space where word embeddings from two or more languages are integrated together.
Current state-of-the-art approaches learn these embeddings by aligning two disjoint monolingual vector spaces through an orthogonal transformation which preserves the structure of the monolingual counterparts.
In this work, we propose to apply an additional transformation after this initial alignment step, which aims to bring the vector representations of a given word and its translations closer to their average. Since this additional transformation is non-orthogonal, it also affects the structure of the monolingual spaces.
We show that our approach both improves the integration of the monolingual spaces as well as the quality of the monolingual spaces themselves.
 Furthermore, because our transformation can be applied to an arbitrary number of languages, we are able to effectively obtain a truly multilingual space.
The resulting (monolingual and multilingual) spaces show consistent gains over the current state-of-the-art in standard intrinsic tasks, namely dictionary induction and word similarity, as well as in extrinsic tasks such as cross-lingual hypernym discovery and cross-lingual natural language inference.
\end{abstract}

\maketitle

\section{Introduction}
\label{intro}

A popular research direction in multilingual Natural Language Processing (NLP) consists in learning \textit{mappings} between two or more monolingual word embedding spaces. These mappings, together with the initial monolingual spaces, define a multilingual word embedding space in which words from different languages with a similar meaning are represented as similar vectors. Such multilingual embeddings do not only play a central role in multilingual NLP tasks, but they also provide a natural tool for transferring models that were trained on resource-rich languages (typically English) to other languages, where the availability of annotated data may be more limited.

State-of-the-art models for aligning monolingual word embeddings currently rely on learning an orthogonal mapping from the monolingual embedding of a source language into the embedding of a target language. Somewhat surprisingly, perhaps, this restriction to orthogonal mappings, as opposed to arbitrary linear or even non-linear mappings, has proven crucial to obtain optimal results. The advantages of using orthogonal transformations are two-fold. First, because they are more constrained than arbitrary linear transformations, they can be learned from noisy data in a more robust way. This plays a particularly important role in settings where alignments between monolingual spaces have to be learned from small and/or noisy dictionaries \citep{artetxe-labaka-agirre:2017:Long}, including dictionaries that have been heuristically induced in a purely unsupervised way \citep{artetxe:acl2018,conneau2018word}. Second, orthogonal transformations preserve the distances between the word vectors, which means that the internal structure of the monolingual spaces is not affected by the alignment. Approaches that rely on orthogonal transformations thus have to assume that the word embedding spaces for different languages are approximately isometric \citep{barone2016towards}. However, it has been argued that this assumption is not always satisfied \citep{sogaard2018limitations,yuva2018generalizing,patra2019bilingual}. Moreover, rather than treating the monolingual embeddings as fixed elements, we may intuitively expect that embeddings from different languages may actually be used to improve each other. This idea was exploited by \citet{faruqui2014improving}, who learn linear mappings from two monolingual spaces onto a new, shared, multilingual space. They found that the resulting changes to the internal structure of the monolingual spaces can indeed bring benefits. In multilingual evaluation tasks, however, their method is outperformed by approaches that rely on orthogonal transformations \citep{artetxe2016learning}.

While the emphasis has shifted from static word vectors to contextualised language models in recent years, it is worth mentioning that static vectors remain an important case of study. On the one hand, static vectors are still needed in applications where the computational demands of contextualised language models are prohibitive, or where word meaning needs to be captured in the absence of context (e.g.,\ ontology alignment). On the other hand, static vectors can also provide useful prior knowledge when training contextualised models such as mBERT~\citep{devlin2019bert}. In particular,  
\citet{artetxe20transferability} show how static cross-lingual embeddings can be exploited for zero-shot multilingual transfer of contextualised models. 

In this article, we propose a simple method which combines the advantages of orthogonal transformations with the potential benefit of allowing monolingual spaces to affect each other's internal structure. Specifically, we first align the given monolingual spaces by learning an orthogonal transformation using an existing state-of-the-art method. Subsequently, we aim to reduce any remaining discrepancies by trying to find the middle ground between the aligned monolingual spaces. Specifically, let $(w,v)$ be an entry from a bilingual dictionary (i.e.,\ $v$ is the translation of $w$), and let $\mathbf{w}$ and $\mathbf{v}$ be the vector representations of $w$ and $v$ in the aligned monolingual spaces. Our aim is to learn linear mappings $\mathbf{M_s}$ and $\mathbf{M_t}$ such that $\mathbf{w}\mathbf{M_s} \approx \mathbf{v}\mathbf{M_t} \approx \frac{\mathbf{v}+\mathbf{w}}{2}$, for each entry $(w,v)$ from a given dictionary. Crucially, because we start from monolingual spaces which are already aligned, applying the mappings $\mathbf{M_s}$ and $\mathbf{M_t}$ can be thought of as a \emph{fine-tuning} step. We will refer to this proposed fine-tuning step as \texttt{\textbf{Meemi}} (\textit{Meeting in the middle})\footnote{Code is available at \url{https://github.com/yeraidm/meemi}. This page will be updated 
with pre-trained models for new languages. 
}. 
Our experimental analysis reveals that this combination of an orthogonal transformation followed by a simple non-orthogonal fine-tuning step consistently, and often substantially outperforms existing approaches in cross-lingual evaluation tasks. We also find that the proposed transformation leads to improvements in the monolingual spaces, which, as already mentioned, is not possible with orthogonal transformations.
This article extends our earlier work in \citet{doval:meemiemnlp2018} in the following ways: 

\begin{enumerate}
    
    \item We introduce a new variant of Meemi, in which the averages that are used to compute the linear transformations are weighted by word frequencies. 
    
    \item We generalize the approach to an arbitrary number of languages, thus allowing us to learn truly multilingual vector spaces.
    
    \item We more thoroughly compare the obtained multilingual models, extending the number of baselines and evaluation tasks. We now also include a more extensive analysis of the results, e.g.\ studying the impact of the size of the bilingual dictionaries in more detail.
    
    \item In the evaluation, we now include two distant languages which do not use the Latin alphabet: Farsi and Russian.
    
\end{enumerate}

\section{Background: Cross-lingual Alignment Methods}
\label{background}

In this article we analyze cross-lingual word embedding models that are based on aligning monolingual vector spaces.
The overall process underpinning these methods is as follows. Given two monolingual corpora, a word vector space is first learned independently for each language. This can be achieved with standard word embedding models such as Word2vec \citep{Mikolovetal:2013}, GloVe \citep{pennington2014glove} or FastText \citep{bojanowski2017enriching}. Second, a linear alignment strategy is used to map the monolingual embeddings to a common bilingual vector space. 
It is worth mentioning that we do not require parallel or comparable corpora to build our multilingual models as in the case of \citet{zennaki2019neural} or \citet{vulic2016bilingual}.

These linear transformations are learned from a supervision signal in the form of a bilingual dictionary (although some methods can also deal with dictionaries that are automatically generated as part of the alignment process; see below). 
This approach was popularized by \citet{mikolov2013exploiting}.  
Specifically, they proposed to learn a matrix $\mathbf{W}$ which minimizes the following objective:
\begin{equation}\label{eqBasicObjectiveAlignment}
\sum_{i=1}^n \| \mathbf{x_i}\mathbf{W} - \mathbf{z_i} \|^2
\end{equation}
where we write $\mathbf{x_i}$ for the vector representation of some word $x_i$ in the source language and $\mathbf{z_i}$ is the vector representation of the translation $z_i$ of $w_i$ in the target language. This optimization problem corresponds to a standard least-squares regression problem, whose exact solution can be efficiently computed (although \citet{mikolov2013exploiting} do not use this method). Note that this approach relies on a bilingual dictionary containing the training pairs $(x_1,z_1),...,(x_n,z_n)$. However, once the matrix $\mathbf{W}$ has been learned, for any word $w$ in the source language, we can use $\mathbf{x}\mathbf{W}$ as a prediction of the vector representation of the translation of $w$. In particular, to predict which word in the target language is the most likely translation of the word $w$ from the source language, we can then simply take the word $z$ whose vector $\mathbf{z}$ is closest to the prediction $\mathbf{x}\mathbf{W}$.

The restriction to linear mappings might intuitively seem overly strict. However, it was found that higher-quality alignments can be found by being even more restrictive. In particular, \citet{xing2015normalized} suggested to normalize the word vectors in the monolingual spaces, and restrict the matrix $\mathbf{W}$ to an orthogonal matrix (i.e.,\ imposing the constraint that  $\mathbf{W}\mathbf{W}^T=\mathbf{1}$). Under this restriction, the optimization problem \eqref{eqBasicObjectiveAlignment} is known as the orthogonal Procrustes problem, whose exact solution can still be computed efficiently. Another approach was taken by \citet{faruqui2014improving}, who proposed to learn linear transformations $\mathbf{W_s}$ and $\mathbf{W_t}$, which respectively map vectors from the source and target language word embeddings onto a shared vector space. They used Canonical Correlation Analysis to find the transformations $\mathbf{W_s}$ and $\mathbf{W_t}$ which minimize the dimension-wise covariance between $\mathbf{X}\mathbf{W_s}$ and $\mathbf{Z}\mathbf{W_t}$, where $\mathbf{X}$ is a matrix whose rows are $\mathbf{x_1},...,\mathbf{x_n}$ and similarly $\mathbf{Z}$ is a matrix whose rows are $\mathbf{z_1},...,\mathbf{z_n}$. Note that while the aim of \citet{xing2015normalized} is to avoid making changes to the cosine similarities between word vectors from the same language, \citet{faruqui2014improving} specifically want to take into account information from the other language with the aim of improving the monolingual embeddings themselves. \citet{artetxe2016learning} propose a model which combines ideas from \citet{xing2015normalized} and \citet{faruqui2014improving}. Specifically, they use the formulation in \eqref{eqBasicObjectiveAlignment} with the constraint that $\mathbf{W}$ be orthogonal, as in \citet{xing2015normalized}, but they also apply a preprocessing strategy called mean centering which is closely related to the model from \citet{faruqui2014improving}. 
On top of this, in \citet{artetxe2018generalizing} they propose a multi-step framework in which they experiment with several pre-processing and post-processing strategies. These include whitening (which involves applying a linear transformation to the word vectors such that their covariance matrix is the identity matrix), re-weighting each coordinate according to its cross-correlation (which means that the relative importance of those coordinates with the strongest agreement between both languages is increased), de-whitening (i.e.,\ inverting the whitening step to restore the original covariances), and a dimensionality reduction step, which is seen as an extreme form of re-weighting (i.e.,\ those coordinates with the least agreement across both languages are simply dropped).
They also consider the possibility of using orthogonal mappings from both embedding spaces into a shared space, rather than mapping one embedding space onto the other, where the objective is based on maximizing cross-covariance.
This route is also followed by \citet{yuva2018generalizing}.
Other approaches that have been proposed for aligning monolingual word embedding spaces include models which replace \eqref{eqBasicObjectiveAlignment} with a max-margin objective \citep{lazaridou2015hubness} and models which rely on neural networks to learn non-linear transformations \citep{lu2015deep}.

A central requirement of the aforementioned methods is that they need a sufficiently large bilingual dictionary. Several approaches have been proposed to address this limitation, showing that high-quality results can be obtained in a purely unsupervised way. For instance, \citet{artetxe-labaka-agirre:2017:Long} propose a method that can work with a small synthetic seed dictionary, e.g.,\ only containing pairs of identical numerals (1,1), (2,2), (3,3), etc. To this end, they alternatingly use the current dictionary to learn a corresponding orthogonal transformation and then use the learned cross-lingual embedding to improve the synthetic dictionary. This improved dictionary is constructed by assuming that the translation of a given word $w$ is the nearest neighbor of $\mathbf{x}\mathbf{W}$ among all words from the target language. This approach was subsequently improved in \citet{artetxe:acl2018}, where state-of-the-art results were  obtained without even assuming the availability of a synthetic seed dictionary. The key idea underlying their approach, called VecMap, is to initialize the seed dictionary in a fully unsupervised way based on the idea that the histogram of similarity scores between a given word $w$ and the other words from the source language should be similar to the histogram of similarity scores between its translation $z$ and the other words from the target language. Another approach which aims to learn bilingual word embeddings in a fully unsupervised way, called MUSE, is proposed in \citet{conneau2018word}. The main difference with VecMap lies in how the initial seed dictionary is learned. For this purpose, MUSE relies on adversarial training \citep{goodfellow2014generative}, similar as in earlier models \citep{barone2016towards,zhang2017adversarial} but using a simpler formulation, based on the model in \eqref{eqBasicObjectiveAlignment} with the orthogonality constraint on $\mathbf{W}$. The main intuition is to choose $\mathbf{W}$ such that it is difficult for a classifier to distinguish between word vectors $\mathbf{z}$ sampled from the target word embedding and vectors $\mathbf{x}\mathbf{W}$, with $\mathbf{x}$ sampled from the source word embedding. There have been other approaches to create this initial bilingual dictionary without supervision via adversarial training \citep{zhang-etal-2017-earth,hoshen-wolf-2018-non,xu2018crosslingual} or stochastic processes \citep{alvarez-melis-jaakkola-2018-gromov}, but their performance has not generally surpassed existing methods \citep{artetxe:acl2018,glavas-etal-2019-properly}. For a more comprehensive summary of existing methods, please refer to \citet{ruder2019survey}.

In this work, we make use of the three mentioned variants of VecMap, namely the supervised implementation based on the multi-step framework from \citet{artetxe2018generalizing}, which will be referred to as VecMap\textsubscript{multistep}, the orthogonal method (VecMap\textsubscript{ortho})~\citep{artetxe2016learning} and its unsupervised version (VecMap\textsubscript{uns}) \citep{artetxe:acl2018}. Similarly, we will consider the supervised and unsupervised variants of MUSE (MUSE and MUSE\textsubscript{uns}, respectively) \citep{conneau2018word}.
In the next section we present our proposed post-processing method based on an unconstrained linear transformation to improve the results of the previous methods.\footnote{Other works have also shown how the orthogonal constrain can be relaxed \citep{joulin2018loss} when training with a specific bilingual dictionary induction objective, but this has been shown not to be optimal for other tasks \citep{glavas-etal-2019-properly}.}

\section{Fine-tuning Cross-lingual Embeddings by Meeting in the Middle}
\label{meemi}

\begin{figure}[t]
\centering
\includegraphics[width=.8\columnwidth]{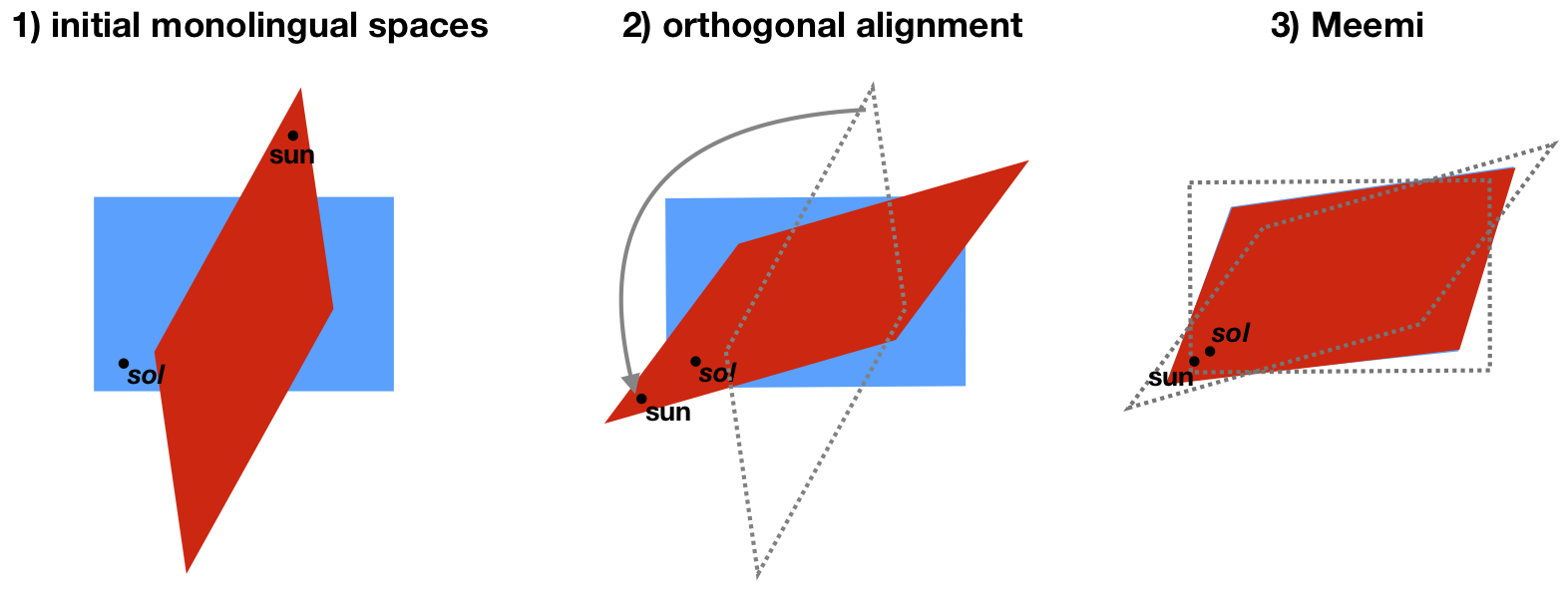}
\caption{Step by step integration of two monolingual embedding spaces: (1) obtaining isolated monolingual spaces, (2) aligning these spaces through an orthogonal linear transformation, and (3) map both spaces using an unconstrained linear transformation learned on the averages of translation pairs.}
\label{fig:cl_diagram}
\end{figure}

After the initial alignment of the monolingual spaces, we propose to apply a post-processing step which aims to bring the two monolingual spaces closer together by lifting the orthogonality constraint.
To this end, we learn an unconstrained linear transformation that maps word vectors from one space onto the average of that word vector and the vector representation of its translation (according to a given bilingual dictionary). 
This approach, which we call Meemi (Meeting in the middle) is illustrated in  Figure~\ref{fig:cl_diagram}. In particular, the figure illustrates the two-step nature, where we first learn an orthogonal transformation (using VecMap or MUSE), which aligns the two monolingual spaces as much as possible without changing their internal structure. Then, our approach aims to find a middle ground between the two resulting monolingual spaces. This involves applying a non-orthogonal transformation to both monolingual spaces. 

By averaging between the representations obtained from different languages, we hypothesize that the impact of language-specific phenomena and corpus specific biases will be reduced whereas its core semantic features will become more dominant.
However, because we start from aligned spaces, the changes which are made by this transformation are relatively small. Our transformation is thus intuitively fine-tuning the usual orthogonal transformation, rather than replacing it.
Note that this approach can naturally be applied to more than two monolingual spaces (Section \ref{multilingual}). First, however, we will consider the standard bilingual case.

\subsection{Bilingual models} 
Let $D$ be the given bilingual dictionary, encoded as a set of word pairs $(w,w')$. Using the pairs in $D$ as training data, we learn a linear mapping $\mathbf{X}$ such that $\mathbf{w} \mathbf{X} \approx \frac{\mathbf{w}+\mathbf{w'}}{2}$ for all $(w,w')\in D$, where we write $\mathbf{w}$ for the vector representation of word $w$ in the given (aligned) monolingual space. This mapping $\mathbf{X}$ can then be used to predict the averages for words outside the given dictionary. 
To find the mapping $\mathbf{X}$, we solve the following least squares linear regression problem:
\begin{equation}\label{eqRegressionFormula}
    E=\sum_{(w,w') \in D} \left\|\mathbf{w} \mathbf{X}-\frac{\mathbf{w}+\mathbf{w'}}{2}\right\|^2
\end{equation}
\noindent Similarly, we separately learn a mapping $\mathbf{X'}$ such that $\mathbf{w'} \mathbf{X'} \approx \frac{\mathbf{w}+\mathbf{w'}}{2}$.

It is worth mentioning that we had also experimented with non-linear mappings before arriving at the present formulation. However, multilayer perceptrons paired with different regularization terms to avoid overfitting, such as penalizing mappings that deviated excessively from the identity mapping, obtained lower performance figures, which led us to discard this path at the moment.

We also consider a weighted variant of Meemi where the linear model is trained on weighted averages based on word frequency.
Specifically, let $f_{w}$ be the occurrence count of word $w$ in the corresponding monolingual corpus, then $\frac{\mathbf{w}+\mathbf{w'}}{2}$ is replaced by:
\begin{equation}\label{eqMultiWeightedAvgFormula}
\frac{f_{w} \mathbf{w} + f_{w'} \mathbf{w'}}{f_{w} + f_{w'}}
\end{equation}
The intuition behind this weighted model is that the word $w$ might be much more prevalent in the first language than the word $w'$ is in the second language. A clear example is when $w=w'$, which may be the case, among others, if $w$ is a named entity. For instance, suppose that $w$ is the name of a Spanish city. Then, we may expect to see more occurrences of $w$ in a Spanish corpus than in an English corpus. In such cases, it may be beneficial to consider the word vector obtained from the Spanish corpus to be of higher quality, and thus give more weight to it in the average.

We will write Meemi ($M$) to refer to the model obtained by applying Meemi after the base method $M$, where $M$ may be any variant of VecMap or MUSE. Similarly, we will write Meemi\textsubscript{w} ($M$) in those cases where the weighted version of Meemi was used. 

\subsection{Multilingual models} 
\label{multilingual}

To apply Meemi in a multilingual setting,
 we exploit the fact that bilingual orthogonal methods such as VecMap (without re-weighting) and MUSE do not modify the target monolingual space but only apply an orthogonal transformation to the source.
Hence, by simply applying this method to multiple language pairs while fixing the target language (i.e., for languages $l_{1}, l_{2}, ..., l_{n}$, we construct pairs of the form $(l_{i}, l_{n})$ with $i \in \{1,...,n-1\}$), we can obtain a multilingual space in which all of the corresponding monolingual models are aligned with, or mapped onto, the same target embedding space. 
Note, however, that if we applied a re-weighting strategy, as suggested in~\citet{artetxe2018generalizing} for VecMap, the target space would no longer remain fixed for all source languages and would instead change depending on the source in each case.
While most previous work has been limited to bilingual settings, multilingual models involving more than two languages have already been studied by~\citet{ammar2016massively}, who used an approach based on Canonical Correlation Analysis.
As in our approach, they also fix one specific language as the reference language. 

Formally, let $D$ be the given multilingual dictionary, encoded as a set of tuples $(w_1,w_2,...,w_n)$, where $n$ is the number of languages. 
Using the tuples in $D$ as training data, we learn a linear mapping $\mathbf{X_i}$ for each language, such that $\mathbf{w_i}\mathbf{X_i} \approx \frac{\mathbf{w_1}+...+\mathbf{w_n}} {n}$ for all $(w_1,...,w_n)\in D$. This mapping $\mathbf{X_i}$ can then be used to predict the averages for words in the $i$th language outside the given dictionary. 
To find the mappings $\mathbf{X_i}$, we solve the following least squares linear regression problem for each language:
\begin{equation}\label{eqMultiRegressionFormula}
    E_{\textit{multi}}=\sum_{(w_1,...,w_n) \in D} \left\|\mathbf{w_i}\mathbf{X_i}- \frac{\mathbf{w_1}+...+\mathbf{w_n}} {n}\right\|^2
\end{equation}
Note that while a weighted variant of this model can straightforwardly be formulated, we will not consider this in the experiments.

\section{Experimental Setting}

In this section we explain the common training settings for all experiments. 
First, the monolingual corpora that were used, as well as other training details that pertain to the initial monolingual embeddings, are discussed in Section \ref{corpora}. Then, in Section \ref{dictionaries} we explain which bilingual and multilingual dictionaries were used as supervision signals. Finally, all compared systems are listed in Section \ref{comparison-systems}.

\subsection{Corpora and monolingual embeddings}
\label{corpora}

Instead of using comparable corpora such as Wikipedia, as in much of the previous work~\citep{artetxe-labaka-agirre:2017:Long,conneau2018word}, we make use of independent corpora extracted from the web. This represents a more realistic setting where alignments are harder to obtain, as already noted by \citet{artetxe:acl2018}. 
For English we use the 3B-word UMBC WebBase Corpus~\citep{han2013umbc}, containing over 3 billion words. 
For Spanish we used the Spanish Billion Words Corpus~\citep{cardellinoSBWCE}, consisting of over a billion words. 
For Italian and German, we use the itWaC and sdeWaC corpora from the WaCky project~\citep{baroni2009wacky}, containing 2 and 0.8 billion words, respectively.\footnote{The same English, Spanish, and Italian corpora are used as input corpora for the hypernym discovery SemEval task (Section \ref{hypernym_discovery}). }  
For Finnish and Russian, we use their corresponding Common Crawl monolingual corpora from the Machine Translation of News Shared Task 2016,\footnote{\url{http://www.statmt.org/wmt16/translation-task.html}} composed of 2.8B and 1.1B words, respectively. 
Finally, for Farsi we leverage the newswire Hamshahri corpus \citep{aleahmad2009hamshahri}, composed of almost 200M words.

In a preprocessing step, all corpora were tokenized using the Stanford tokenizer~\citep{manning2014stanford} and lowercased. Then we trained FastText word embeddings \citep{bojanowski2017enriching} on the preprocessed corpora for each language. The dimensionality of the vectors was set to 300, using the default values for the remaining hyperparameters. 

In our experiments we consider, first, 6 Indo-European languages, of which Spanish and Italian are Romance, English and German are Germanic, Russian is Slavic, and Farsi is Iranian. Second, we also include experiments for Finnish, which is a Uralic Finnic language~\citep{wals}. 
Finally, we have also included a set of exclusively distant languages: Arabic and Hebrew, both of them Semitic Afro-Asiatic; Finnic Uralic Estonian, Slavic Indo-European Polish, and Sino-Tibetan Chinese. For this latter set of languages, we use the pretrained monolingual embeddings available from the FastText website,\footnote{https://fasttext.cc/docs/en/crawl-vectors.html} obtained from Common Crawl and Wikipedia.
Since we could not access the source corpora for these monolingual embeddings, we could not gather frequency information and therefore we only tested the default variant of Meemi (i.e., not weighted). Furthermore, for the multilingual version of Meemi (see Section \ref{multilingual}), we consider those languages for which we train the corresponding monolingual embeddings: English, Spanish, Italian, German, Russian, Farsi and Finnish. 

\subsection{Training dictionaries}
\label{dictionaries}

We use the training dictionaries provided by \citet{conneau2018word} as supervision. These bilingual dictionaries were compiled using the internal translation tools from Facebook. To make the experiments comparable across languages, we randomly extracted 8,000 training pairs for all language pairs considered, as this is the size of the smallest available dictionary. For completeness we also present results for fully unsupervised systems (see the following section), which do not take advantage of any dictionaries. 

\subsection{Compared systems}
\label{comparison-systems}

We have trained both bilingual and multilingual models involving up to seven languages.
In the bilingual case, we consider the supervised and unsupervised variants of VecMap and MUSE to obtain the base alignments and then apply \emph{plain} Meemi and weighted Meemi on the results. For supervised VecMap we compare with its orthogonal version VecMap\textsubscript{ortho} and the multi-step procedure VecMap\textsubscript{multistep}. 
For the multilingual case we follow the procedure described in Section \ref{multilingual} making use of all seven languages considered in the evaluation, i.e., English, Spanish, Italian, German, Finnish, Farsi, and Russian. Note that in the bilingual case all three variants of VecMap can be used, whereas in the multilingual setting we can only use VecMap\textsubscript{ortho}.

\section{Intrinsic evaluation}

In this section we assess the intrinsic performance of our post-processing techniques in cross-lingual (Section \ref{cross-eval}) and monolingual (Section \ref{monoeval}) settings.

\subsection{Cross-lingual performance}
\label{cross-eval}

We evaluate the performance of all compared cross-lingual embedding models on standard purely cross-lingual tasks, namely dictionary induction (Section \ref{induction}) and cross-lingual word similarity (Section \ref{cross-similarity}).

\begin{table}[!ht]
\resizebox{\textwidth}{!}{
\begin{tabular}{|l|rrr|rrr|rrr|}
\hline
\multicolumn{1}{|c|}{\multirow{2}{*}{\textbf{Model}}}                               & \multicolumn{3}{c|}{\textbf{English-Spanish}}                                                                    & \multicolumn{3}{c|}{\textbf{English-Italian}}                                                                     & \multicolumn{3}{c|}{\textbf{English-German}}                                                                    \\ \cline{2-10} 
\multicolumn{1}{|c|}{}                                                              & \multicolumn{1}{c|}{\textbf{$P@1$}} & \multicolumn{1}{c|}{\textbf{$P@5$}} & \multicolumn{1}{c|}{\textbf{$P@10$}} & \multicolumn{1}{c|}{\textbf{$P@1$}} & \multicolumn{1}{c|}{\textbf{$P@5$}} & \multicolumn{1}{c|}{\textbf{$P@10$}} & \multicolumn{1}{c|}{\textbf{$P@1$}} & \multicolumn{1}{c|}{\textbf{$P@5$}} & \multicolumn{1}{c|}{\textbf{$P@10$}} \\ \hline
VecMap\textsubscript{uns}                                         & \textbf{34.8}                                & 60.6                                & \multicolumn{1}{r|}{67.0}            & 31.4                                & 53.7                                & \multicolumn{1}{r|}{60.7}            & 23.2                                & 42.7                                & 50.2                                \\
MUSE\textsubscript{uns}                                        & 31.4                                & 51.2                                & \multicolumn{1}{r|}{57.7}            & 31.4                                & 51.2                                & \multicolumn{1}{r|}{57.7}            & 20.8                                & 38.7                                & 46.6                                \\
\hline
VecMap\textsubscript{ortho}                                         & 32.6                                & 58.1                                & \multicolumn{1}{r|}{65.8}            & 32.9                                & 56.5                                & \multicolumn{1}{r|}{63.4}            & 22.8                                & 42.8                                & 50.4                                \\
Meemi (VecMap\textsubscript{ortho})                                 & 33.9                                & 60.7                                & \multicolumn{1}{r|}{67.4}            & \textbf{33.8}                                & 58.8                                & \multicolumn{1}{r|}{65.6}            & 23.7                                & 45.0                                & 52.9                                \\
Meemi\textsubscript{w} (VecMap\textsubscript{ortho}) & 33.4                                & 60.9                                & \multicolumn{1}{r|}{67.4}            & 33.1                                & 58.5                                & \multicolumn{1}{r|}{66.3}            & 22.9                                & 44.3                                & 52.5                                \\
Meemi-multi (VecMap\textsubscript{ortho}) & 33.4 & 60.9 & 67.1 & 33.7 & 58.1 & 65.5 & 23.0 & 44.5 & 52.8 \\
\hline

VecMap\textsubscript{multistep}                                                                             & 33.8                                & 60.7                                & \multicolumn{1}{r|}{\textbf{68.4}}            & 33.7                                & 58.9                                & \multicolumn{1}{r|}{66.5}            & \textbf{24.1}                                & 45.3                                & \textbf{53.6}                                \\
Meemi (VecMap\textsubscript{multistep})                                                                     & 33.8                                & \textbf{61.4}                                & \multicolumn{1}{r|}{\textbf{68.4}}            & 33.7                                & \textbf{59.0}                                & \multicolumn{1}{r|}{\textbf{66.8}}            & 23.4                                & \textbf{45.7}                                & \textbf{53.6}                                \\

Meemi\textsubscript{w} (VecMap\textsubscript{multistep})                                     & 33.2                                & 60.9                                & \multicolumn{1}{r|}{68.1}            & 32.5                                & 58.2                                & \multicolumn{1}{r|}{66.2}            & 22.8                                & 44.8                                & 53.1                                \\
\hline

MUSE                                                                               & 32.5                                & 58.2                                & \multicolumn{1}{r|}{65.9}            & 32.5                                & 56.0                                & \multicolumn{1}{r|}{63.2}            & 22.4                                & 40.9                                & 48.9                                \\
Meemi (MUSE)                                                                       & 33.9                                & 60.7                                & \multicolumn{1}{r|}{\textbf{68.4}}            & \textbf{33.8}                                & 58.4                                & \multicolumn{1}{r|}{65.6}            & 23.7                                & 45.3                                & 52.3                                \\
Meemi\textsubscript{w} (MUSE)                                       & 33.3                                & 61.2                                & \multicolumn{1}{r|}{68.2}            & 33.0                                & 58.8                                & \multicolumn{1}{r|}{65.3}            & 22.8                                & 44.4                                & 52.3                                \\ \hline \hline

\multicolumn{1}{|c|}{\multirow{2}{*}{\textbf{Model}}}                                                              & \multicolumn{3}{c|}{\textbf{English-Finnish}}                                                                    & \multicolumn{3}{c|}{\textbf{English-Farsi}}                                                                       & \multicolumn{3}{c|}{\textbf{English-Russian}}                                                                   \\ \cline{2-10} 
\multicolumn{1}{|l|}{}                                                              & \multicolumn{1}{c|}{\textbf{$P@1$}} & \multicolumn{1}{c|}{\textbf{$P@5$}} & \multicolumn{1}{c|}{\textbf{$P@10$}} & \multicolumn{1}{c|}{\textbf{$P@1$}} & \multicolumn{1}{c|}{\textbf{$P@5$}} & \multicolumn{1}{c|}{\textbf{$P@10$}} & \multicolumn{1}{c|}{\textbf{$P@1$}} & \multicolumn{1}{c|}{\textbf{$P@5$}} & \multicolumn{1}{c|}{\textbf{$P@10$}} \\ \hline
VecMap\textsubscript{uns}                                         & 0.1                                 & 0.5                                 & 0.7                                  & 19.7                                & 34.6                                & 40.4                                 & 13.8                                & 30.9                                & 38.6                                \\
MUSE\textsubscript{uns}                                           & 23.7                                & 45.0                                & 52.9                                 & 18.1                                & 32.8                                & 37.8                                 & 14.4                                 & 31.2                                 & 38.5                                 \\
\hline

VecMap\textsubscript{ortho}                                         & 22.1                                & 44.5                                & 52.9                                 & 18.5                                & 33.6                                & 40.5                                 & 15.6                                & 35.5                                & 44.2                                \\
Meemi (VecMap\textsubscript{ortho})                                 & \textbf{24.8}                                & 48.9                                & 57.7                                 & 20.0                                & 37.1                                & \textbf{43.8}                                 & 19.0                                & 40.5                                & 49.9                                \\
Meemi\textsubscript{w} (VecMap\textsubscript{ortho}) & 22.6                                & 48.3                                & 56.5                                 & 19.8                                & 35.2                                & 41.6                                 & 17.4                                & 39.9                                & 49.4                                \\
Meemi-multi (VecMap\textsubscript{ortho}) & 23.1 & 48.3 & 57.2 & 21.0 & 37.9 & 44.4 & 18.8 & 41.7 & 50.5 \\
\hline

VecMap\textsubscript{multistep}                                                                             & 22.5                                & 48.4                                & 57.5                                 & 20.8                                & 36.1                                & 43.4                                 & 18.2                                & 40.2                                & 49.5                                \\
Meemi (VecMap\textsubscript{multistep})                                                                     & 24.0                                & \textbf{50.8}                                & \textbf{58.9}                                 & 20.0                                & 36.9                                & 42.4                                 & \textbf{19.3}                                & \textbf{41.5}                                & \textbf{50.6}                                \\
Meemi\textsubscript{w} (VecMap\textsubscript{multistep})                                     & 21.6                                & 48.3                                & 57.2                                 & \textbf{21.5}                                & \textbf{38.5}                                & 43.7                                 & 17.4                                & 40.9                                & 49.7                                \\
\hline

MUSE                                                                               & 20.0                                & 40.1                                & 48.3                                 & 17.4                                & 31.6                                & 37.6                                 & 15.5                                & 35.6                                & 44.1                                \\
Meemi (MUSE)                                                                       & 23.0                                & 46.1                                & 54.0                                 & 19.3                                & 36.0                                & 41.7                                 & 18.7                                & 40.5                                & 49.7                                \\
Meemi\textsubscript{w} (MUSE)                                       & 21.7                                & 46.9                                & 55.0                                 & 19.5                                & 33.8                                & 39.8                                 & 18.1                                & 40.0                                & 49.5   \\      \hline      

\end{tabular}
}
\caption{$P@K$ performance of different cross-lingual embedding models in the bilingual dictionary induction task.}
\label{tab:dictinduction} 
\end{table}

\begin{table}[]
\centering
\resizebox{\textwidth}{!}{%
\begin{tabular}{|l|rrr|rrr|rrr|rrr|rrr|}
\hline
\multicolumn{1}{|c|}{\multirow{2}{*}{\textbf{Model}}} & \multicolumn{3}{c|}{\textbf{English-Arabic}}                                        & \multicolumn{3}{c|}{\textbf{English-Hebrew}}                                        & \multicolumn{3}{c|}{\textbf{English-Estonian}}                                      & \multicolumn{3}{c|}{\textbf{English-Polish}}                                        & \multicolumn{3}{c|}{\textbf{English-Chinese}}                                       \\ \cline{2-16} 
\multicolumn{1}{|c|}{}                                & \multicolumn{1}{c|}{$P@1$} & \multicolumn{1}{c|}{$P@5$} & \multicolumn{1}{c|}{$P@10$} & \multicolumn{1}{c|}{$P@1$} & \multicolumn{1}{c|}{$P@5$} & \multicolumn{1}{c|}{$P@10$} & \multicolumn{1}{c|}{$P@1$} & \multicolumn{1}{c|}{$P@5$} & \multicolumn{1}{c|}{$P@10$} & \multicolumn{1}{c|}{$P@1$} & \multicolumn{1}{c|}{$P@5$} & \multicolumn{1}{c|}{$P@10$} & \multicolumn{1}{c|}{$P@1$} & \multicolumn{1}{c|}{$P@5$} & \multicolumn{1}{c|}{$P@10$} \\ \hline
VecMap\textsubscript{uns}                             & 17.3                     & 34.5                     & 41.1                       & 20.1                     & 35.6                     & 41.1                       & 17.5                     & 33.9                     & 39.5                       & 20.8                     & 40.9                     & 46.7                       & 9.3                      & 17.5                     & 22.1                       \\
MUSE\textsubscript{uns}                               & 0.0                      & 0.0                      & 0.0                        & 0.0                      & 0.0                      & 0.0                        & 8.6                      & 17.9                     & 22.1                       & 10.7                     & 22.8                     & 28.3                       & 0.0                      & 0.0                      & 0.1                        \\ \hline
VecMap\textsubscript{ortho}                           & 18.7                     & 40.8                     & 48.3                       & 19.1                     & 36.0                     & 42.0                       & 19.9                     & 36.2                     & 42.2                       & 23.2                     & 42.3                     & 48.5                       & 20.7                     & 36.7                     & 42.8                       \\
Meemi (VecMap\textsubscript{ortho})                   & 20.0                     & 45.0                     & 52.4                       & 20.5                     & 38.8                     & 45.0                       & 21.3                     & 37.9                     & 43.9                       & \textbf{24.2}                     & \textbf{45.0}                     & 50.1                       & 23.2                     & 40.7                     & 47.4                       \\ \hline
VecMap\textsubscript{multistep}                                                & 20.3                     & 44.8                     & 52.8                       & 20.4                     & 38.6                     & 43.9                       & \textbf{21.5}                     & \textbf{39.5}                     & \textbf{44.6}                       & 23.8                     & \textbf{45.0}                     & \textbf{51.0}                       & 24.1                     & 41.6                     & 48.3                       \\
Meemi (VecMap\textsubscript{multistep})                                        & \textbf{20.8}                     & \textbf{46.6}                     & \textbf{53.5}                       & \textbf{20.9}                     & \textbf{39.2}                     & \textbf{45.2}                       & 21.2                     & 38.3                     & 43.7                       & 23.2                     & 44.8                     & 50.4                       & \textbf{24.2}                     & \textbf{42.7}                     & \textbf{48.6}                       \\ \hline
MUSE                                                  & 16.3                     & 36.5                     & 42.5                       & 15.9                     & 30.1                     & 35.0                       & 14.8                     & 26.7                     & 31.1                       & 18.7                     & 36.8                     & 42.2                       & 16.8                     & 31.2                     & 36.6                       \\
Meemi (MUSE)                                          & 17.3                     & 38.6                     & 45.3                       & 16.6                     & 30.9                     & 36.1                       & 17.9                     & 32.3                     & 36.9                       & 21.1                     & 40.8                     & 46.5                       & 17.1                     & 31.4                     & 37.0                       \\ \hline
\end{tabular}%
}
\caption{Dictionary induction results for distant language pairs using FastText pre-trained monolingual embeddings as input.}
\label{tab:dict_ind_app}
\end{table}

\subsubsection{Bilingual dictionary induction}
\label{induction}

Also referred to as word translation, this task consists in automatically retrieving the word translations in a target language for words in a source language.
Acting on the corresponding cross-lingual embedding space which integrates the two (or more) languages of a particular test case, we obtain the nearest neighbors to the source word in the target language as our translation candidates.
The performance is measured with precision at $k$ ($P@k$), defined as the proportion of test instances where the correct translation candidate for a given source word was among the $k$ highest ranked candidates. 
The nearest neighbors ranking is obtained by using cosine similarity as the scoring function. For this evaluation we use the corresponding test dictionaries released by \citet{conneau2018word}.

We show the results attained by a wide array of models in Tables~\ref{tab:dictinduction} and~\ref{tab:dict_ind_app}, where we can observe that the best figures are generally obtained by Meemi over the bilingual VecMap models. The impact of Meemi is more apparent when used in combination with the orthogonal base models, with improvements over the multi-step version of VecMap as well in most languages. 
These improvements are statistically significant at the 0.05 level across all language pairs, using paired t-tests.
On the other hand, using the weighted version of Meemi (i.e., Meemi\textsubscript{w} in Table~\ref{tab:dictinduction}) does not seem to be particularly beneficial on this task, with the only exception of English-Farsi. In general, the performance of unsupervised models (i.e., VecMap\textsubscript{uns} and MUSE\textsubscript{uns}) is competitive in closely-related languages such as English-Spanish or English-German but they considerably under-perform for distant languages, especially English-Finnish and English-Russian. 
We have double-checked the anomalous results for English-Finnish, and they appear to be correct under our current testing framework after multiple runs.
Finally, the results obtained by the multilingual model that includes all seven languages considered, i.e.,\ Meemi-multi (VecMap\textsubscript{ortho}) in Table~\ref{tab:dictinduction}, improve over the base orthogonal model, but they do not improve over the results of our bilingual model.  
We further discuss the impact of adding languages to the multilingual model in Section~\ref{multi-eval}.

\subsubsection{Cross-lingual word similarity}
\label{cross-similarity}

Cross-lingual word similarity constitutes a straightforward benchmark to test the quality of bilingual embeddings. In this case, and in contrast to monolingual similarity, words in a given pair \textit{(a,b)} belong to different languages, e.g., \textit{a} belonging to English and \textit{b} to Farsi. For this task we make use of the SemEval-17 multilingual similarity benchmark \citep{semeval2017similarity}, considering the four cross-lingual datasets that include English as target language in particular, but discarding multi-word expressions. Also, we use the Multi-SimLex dataset published by~\citet{vulic2020multisimlex} for our experiments on the set of exclusively distant languages: Arabic, Hebrew, Estonian, Polish, and Chinese.
Performance is computed in terms of Pearson and Spearman correlation with respect to the gold standard.

Tables \ref{tab:cross-sim} and~\ref{tab:wsim_app} show the results of the different embeddings models in the cross-lingual word similarity task. Except in a few cases for the VecMap\textsubscript{multistep} model, our Meemi transformation proves superior to the base models (at the 0.05 level for paired t-tests over all language pairs), and to all their unsupervised variants. For distant languages, where the results are lower overall, our Meemi transformation proves useful, generally outperforming the best VecMap models. 
Similarly as in the bilingual dictionary induction task, the weighted version of Meemi proves robust only on English-Farsi (Table~\ref{tab:wsim_app}), which suggests that this weighting scheme is most useful for distant languages, as in this case the Farsi monolingual space (which is learned from a smaller corpus and hence, as we will see in the next section, has a lower quality) gets closer to the English monolingual space. As far as the multilingual model is concerned, it proves beneficial in all cases with respect to the orthogonal version of VecMap, as well as compared to the bilingual variant of Meemi.

\begin{table}[!t]
\resizebox{\textwidth}{!}{
\begin{tabular}{|l|rr|rr|rr|rr|}
\hline
\multicolumn{1}{|c|}{\multirow{2}{*}{\textbf{Model}}}                                        & \multicolumn{2}{c|}{\textbf{EN-ES}}                    & \multicolumn{2}{c|}{\textbf{EN-IT}}                   & \multicolumn{2}{c|}{\textbf{EN-DE}}                    & \multicolumn{2}{c|}{\textbf{EN-FA}}                    \\ \cline{2-9} 
\multicolumn{1}{|c|}{}                                                              & \multicolumn{1}{c|}{\textbf{$r$}} & \multicolumn{1}{c|}{\textbf{$\rho$}} & \multicolumn{1}{c|}{\textbf{$r$}} & \multicolumn{1}{c|}{\textbf{$\rho$}} & \multicolumn{1}{c|}{\textbf{$r$}} & \multicolumn{1}{c|}{\textbf{$\rho$}} & \multicolumn{1}{c|}{\textbf{$r$}} & \multicolumn{1}{c|}{\textbf{$\rho$}} \\ \hline
VecMap\textsubscript{uns}                                         & 71.1                     & 70.5                        & 69.2                     & 68.8                        & 70.9                     & 70.4                        & 35.7                     & 33.4                       \\
MUSE\textsubscript{uns}                                           & 71.7                     & 71.6                        & 69.4                     & 69.4                        & 70.3                     & 70.0                        & 29.6                     & 23.8                       \\
\hline
VecMap\textsubscript{ortho} & 71.6                     & 71.6                        & 70.2                     & 70.1                        & 70.9                     & 70.7                        & 29.2                     & 23.7                       \\
Meemi (VecMap\textsubscript{ortho})                                 & 72.3                     & 72.0                        & 71.2                     & 70.7                        & 72.5                     & 72.1                        & 35.3                     & 31.6                       \\
Meemi\textsubscript{w} (VecMap\textsubscript{ortho}) & 72.1                     & 72.0                        & 70.0                     & 69.7                        & 70.5                     & 70.2                        & 34.2                     & 30.2                       \\
Meemi-multi (VecMap\textsubscript{ortho})                           & \textbf{73.9}                     & \textbf{73.4}                       & \textbf{71.6}                     & 71.0                        & 72.5                     & 72.2                        & 39.6                     & 37.2                       \\
\hline
VecMap\textsubscript{multistep}                                                                             & 72.8                     & 72.4                        & \textbf{71.6}                     & \textbf{71.2}                        & \textbf{72.7}                     & 72.2                        & 36.5                     & 31.7                       \\
Meemi (VecMap\textsubscript{multistep}  )                                                                     & 72.1                     & 71.5                        & 71.1                     & 70.9                        & 72.6                     & \textbf{72.3}                        & \textbf{40.4}                     & 39.0                       \\
Meemi\textsubscript{w} (VecMap\textsubscript{multistep}  )                                     & 71.5                     & 71.2                        & 69.7                     & 69.8                        & 70.3                     & 70.3                        & 39.6                     & \textbf{40.8}                       \\
\hline
MUSE                                                                               & 71.9                     & 71.9                        & 70.4                     & 70.4                        & 70.5                     & 70.2                        & 29.7                     & 23.9                       \\
Meemi (MUSE)                                                                       & 72.5                     & 72.3                        & 71.5                     & 71.1                        & 72.5                     & 72.1                        & 36.4                     & 33.0                       \\
Meemi\textsubscript{w} (MUSE)                                       & 72.3                     & 72.2                        & 70.4                     & 70.0                        & 70.5                     & 70.4                        & 33.6                     & 28.9           \\ \hline           
\end{tabular}
}
\caption{Cross-lingual word similarity results in terms of Pearson ($r$) and Spearman ($\rho$) correlation. Languages codes: English-EN, Spanish-ES, Italian-IT, German-DE, and Farsi-FA.}
\label{tab:cross-sim}
\end{table}

\begin{table}[]
\centering
\resizebox{\textwidth}{!}{%
\begin{tabular}{|l|rr|rr|rr|rr|rr|}
\hline
\multicolumn{1}{|c|}{\multirow{2}{*}{\textbf{Model}}} & \multicolumn{2}{c|}{\textbf{EN-AR}}           & \multicolumn{2}{c|}{\textbf{EN-HE}}           & \multicolumn{2}{c|}{\textbf{EN-ET}}         & \multicolumn{2}{c|}{\textbf{EN-PL}}           & \multicolumn{2}{c|}{\textbf{EN-ZH}}          \\ \cline{2-11} 
\multicolumn{1}{|c|}{}                                & \multicolumn{1}{c|}{$r$} & \multicolumn{1}{c|}{$\rho$} & \multicolumn{1}{c|}{$r$} & \multicolumn{1}{c|}{$\rho$} & \multicolumn{1}{c|}{$r$} & \multicolumn{1}{c|}{$\rho$} & \multicolumn{1}{c|}{$r$} & \multicolumn{1}{c|}{$\rho$} & \multicolumn{1}{c|}{$r$} & \multicolumn{1}{c|}{$\rho$} \\ \hline
VecMap\textsubscript{uns}              & 17.4                    & 34.5                       & 41.2                    & 20.2                       & 35.6                    & 41.2                       & 17.5                    & 33.9                       & 39.6                    & 20.8                       \\
MUSE\textsubscript{uns}                & 0.0                     & 0.0                        & 0.0                     & 0.0                        & 0.0                     & 0.0                        & 8.7                     & 17.9                       & 22.1                    & 10.7                       \\ \hline
VecMap\textsubscript{ortho}            & 18.7                    & 40.8                       & 48.3                    & 19.1                       & 36.0                    & 42.0                       & 19.9                    & 36.2                       & 42.2                    & 23.2                       \\
Meemi (VecMap\textsubscript{ortho})    & 20.0                    & 45.0                       & 52.4                    & 20.5                       & 38.8                    & 45.0                       & 21.3                    & 37.9                       & 43.9                    & \textbf{24.2}                       \\ \hline
VecMap\textsubscript{multistep}                                                & 20.3                    & 44.8                       & 52.8                    & 20.4                       & 38.6                    & 43.9                       & \textbf{21.5}                    & \textbf{39.5}                       & \textbf{44.6}                    & 23.8                       \\
Meemi (VecMap\textsubscript{multistep})                                        & \textbf{20.8}                    & \textbf{46.6}                       & \textbf{53.5}                    & \textbf{20.9}                       & \textbf{39.2}                    & \textbf{45.2}                       & 21.2                    & 38.3                       & 43.7                    & 23.2                       \\ \hline
MUSE                                                  & 16.3                    & 36.5                       & 42.5                    & 15.9                       & 30.1                    & 35.0                       & 14.8                    & 26.7                       & 31.1                    & 18.7                       \\
Meemi (MUSE)                                          & 17.3                    & 38.6                       & 45.3                    & 16.6                       & 30.9                    & 36.1                       & 17.9                    & 32.3                       & 36.9                    & 21.1                       \\ \hline
\end{tabular}%
}
\caption{Cross-lingual word similarity results for distant language pairs using FastText pre-trained monolingual embeddings as input. Language codes: English-EN, Arabic-AR, Hebrew-HE, Estonian-ET, Polish-PL, and Chinese-ZH.}
\label{tab:wsim_app}
\end{table}

As for the results with distant languages in Table \ref{tab:wsim_app} (using pre-trained FastText embeddings), the trend is even more pronounced. Meemi helps improve the performance in all languages for the MUSE and VecMap orthogonal methods, and it also improves the performance of VecMap\textsubscript{multistep} in Arabic, Hebrew and Estonian.

\subsection{Monolingual performance}
\label{monoeval}

One of the advantages of breaking the orthogonality of the transformation is the potential to improve the monolingual quality of the embeddings. To test the difference between the original word embeddings and the embeddings obtained after applying the Meemi transformation, we take monolingual word similarity as a benchmark. Given a word pair, this task consists in assessing the semantic similarity between both words in the pair, in this case from the same language. The evaluation is then performed in terms of Spearman and Pearson correlation with respect to human judgements. In particular, we use the monolingual datasets (English, Spanish, German, and Farsi) from the SemEval-17 task on multilingual word similarity. The results provided by the original monolingual FastText embeddings are also reported as baseline.

Table \ref{tab:mono-sim} shows the results on the monolingual word similarity task. In this task our multilingual model representing seven languages in a single space clearly stands out, obtaining the best overall results for English, Spanish and Italian, and improving over the base VecMap\textsubscript{ortho} model on the rest. With the exception of German, where the multi-step framework of \citet{artetxe2018generalizing} proves most effective, the plain Meemi transformation improves over the base models, for both VecMap and MUSE. 

\begin{table}[!t]
\resizebox{\textwidth}{!}{
\begin{tabular}{|l|rr|rr|rr|rr|rr|}
\hline
\multicolumn{1}{|c|}{\multirow{2}{*}{\textbf{Model}}}                               & \multicolumn{2}{c|}{\textbf{English}}                                    & \multicolumn{2}{c|}{\textbf{Spanish}}                                    & \multicolumn{2}{c|}{\textbf{Italian}}                                    & \multicolumn{2}{c|}{\textbf{German}}                                     & \multicolumn{2}{c|}{\textbf{Farsi}}                                     \\ \cline{2-11} 
\multicolumn{1}{|c|}{}                                                              & \multicolumn{1}{c|}{\textbf{$r$}} & \multicolumn{1}{c|}{\textbf{$\rho$}} & \multicolumn{1}{c|}{\textbf{$r$}} & \multicolumn{1}{c|}{\textbf{$\rho$}} & \multicolumn{1}{c|}{\textbf{$r$}} & \multicolumn{1}{c|}{\textbf{$\rho$}} & \multicolumn{1}{c|}{\textbf{$r$}} & \multicolumn{1}{c|}{\textbf{$\rho$}} & \multicolumn{1}{c|}{\textbf{$r$}} & \multicolumn{1}{c|}{\textbf{$\rho$}} \\ \hline
VecMap\textsubscript{uns}                                         & 72.8                              & 72.3                                 & 70.2                              & 70.4                                 & 67.8                              & 68.1                                 & 70.6                              & 70.2                                 & 23.5                              & 21.1                                \\
MUSE\textsubscript{uns}                                           & 74.2                              & 74.2                                 & 70.5                              & 71.9                                 & 67.4                              & 69.2                                 & 69.8                              & 69.8                                 & 21.1                              & 17.3                                \\
\hline
VecMap\textsubscript{ortho}                                         & 74.1                              & 73.9                                 & 70.0                              & 71.5                                 & 67.2                              & 69.0                                 & 70.1                              & 70.1                                 & 21.1                              & 18.2                                \\
Meemi (VecMap\textsubscript{ortho})                                 & 74.4                              & 73.9                                 & 71.6                              & 72.1                                 & 69.0                              & 69.4                                 & 71.1                              & 70.7                                 & 24.3                              & 22.5                                \\
Meemi\textsubscript{w} (VecMap\textsubscript{ortho}) & 74.4                              & 74.0                                 & 71.8                              & 71.8                                 & 68.2                              & 68.8                                 & 68.8                              & 68.9                                 & \textbf{28.5}                              & \textbf{29.8}                                \\
Meemi-multi (VecMap\textsubscript{ortho})                           & \textbf{75.1}                              & 74.3                                 & \textbf{73.0}                              & \textbf{72.9}                                 & \textbf{70.1}                              & \textbf{70.4}                                 & 70.7                              & 70.7                                 & 27.3                              & 26.0                                \\
\hline
VecMap\textsubscript{multistep}                                                                          & 73.8                              & 73.3                                 & 71.8                              & 72.0                                 & 69.6                              & 69.7                                 & \textbf{71.8}                              &\textbf{71.2}                                 & 24.8                              & 22.2                                \\
Meemi (VecMap\textsubscript{multistep})                                                                     & 73.3                              & 72.6                                 & 71.7                              & 71.6                                 & 69.4                              & 69.8                                 & 71.1                              & 71.0                                 & 27.3                              & 26.2                                \\
Meemi\textsubscript{w} (VecMap\textsubscript{multistep})                                     &     73.5  &       	72.9  &       		70.9  &       	70.6  &       		67.2  &       	68.4  &       		67.0  &       	67.8  &       		27.3  &       	25.6                         \\
\hline
MUSE                                                                               & 74.2                              & 74.2                                 & 70.5                              & 71.9                                 & 67.4                              & 69.2                                 & 69.8                              & 69.8                                 & 21.1                              & 17.3                                \\
Meemi (MUSE)                                                                       & 74.6                              & 74.1                                 & 71.9                              & 72.4                                 & 69.5                              & 69.9                                 & 71.0                              & 70.6                                 & 24.6                              & 22.5                                \\
Meemi\textsubscript{w} (MUSE)                                       & 74.5                              & \textbf{74.4}                                 & 71.7                              & 71.8                                 & 68.5                              & 68.9                                 & 68.3                              & 68.2                                 & 27.0                              & 25.5                                \\
\hline
FastText                                                                           & 72.3                              & 72.4                                 & 69.0                              & 70.2                                 & 66.3                              & 67.5                                 & 71.0                              & 70.3                                 & 24.3                              & 20.6                                \\
\hline
\hline
Human upper bound                                                                   & \textit{89.3}                              & -                                    & \textit{89.0}                              & -                                    &\textit{90.0}                              & -                                    & \textit{91.6}                              & -                                    & \textit{90.6}                              & - \\ \hline                                  
\end{tabular}
}
\caption{Monolingual word similarity results in terms of Pearson ($r$) and Spearman ($\rho$) correlation.}
\label{tab:mono-sim}
\end{table}

\section{Extrinsic evaluation}
\label{extrinsic}

We complement the intrinsic evaluation experiments, which are typically a valuable source for understanding the properties of the vector spaces, with downstream extrinsic cross-lingual tasks. This evaluation is especially necessary in the view that the intrinsic behaviour does not always correlate well with downstream performance \citep{bakarov2018limitations,glavas-etal-2019-properly}. 
In particular, for this extrinsic evaluation we will focus on the following question: how does our post-processing method help alleviate limitations of cross-lingual models that are due to their use of orthogonality constraints? 
In particular, we perform experiments with the orthogonal model of VecMap (i.e., VecMap\textsubscript{ortho}), in combination with the proposed Meemi strategy, both in bilingual and multilingual settings. For the latter case, we considered all six languages, i.e., Spanish, Italian, German, Finnish, Farsi, and Russian, keeping English as the target language.

The tasks considered are cross-lingual hypernym discovery (Section \ref{hypernym_discovery}) and cross-lingual natural language inference (Section \ref{XLI}).

\subsection{Cross-lingual hypernym discovery}
\label{hypernym_discovery}

Hypernymy is an important lexical relation, which, if properly modeled, directly impacts downstream NLP tasks such as semantic search \citep{hoffart2014stics,roller-erk:2016:EMNLP2016}, question answering \citep{prager2008question,yahya2013robust} or textual entailment \citep{geffet2005distributional}. Hypernyms, in addition, are the backbone of taxonomies and lexical ontologies \citep{Yuetal2015}, which are in turn useful for organizing, navigating, and retrieving online content \citep{bordea2016semeval}. We propose to evaluate the quality of a range of cross-lingual vector spaces in the extrinsic task of hypernym discovery, i.e., given an input word (e.g., ``cat''), retrieve or discover its most likely (set of) valid hypernyms (e.g., ``animal'', ``mammal'', ``feline'', and so on). Intuitively, by leveraging a bilingual vector space condensing the semantics of two languages, one of them being English, the need for large amounts of training data in the target language may be reduced.\footnote{Note that this task is more challenging than hypernym detection, which is typically framed as a binary classification problem \citep{upadhyay2018robust}, as the search space is equal to the size of the vocabulary considered for each language.}

The base model is a (cross-lingual) linear transformation trained with hyponym-hypernym pairs \citep{EspinosaEMNLP2016}, which is afterwards used to predict the most likely (set of) hypernyms given a new term. Training and evaluation data come from the SemEval 2018 Shared Task on Hypernym Discovery \citep{semeval2018task9}. Note that current state-of-the-art systems aimed at modeling hypernymy \citep{shwartzetal2016,berniercolborne-barriere:2018:SemEval-2018,held-habash-2019-effectiveness} combine large amounts of annotated data along with language-specific rules and cue phrases such as Hearst Patterns \citep{Hearst:92}, both of which are generally scarcely (if at all) available for languages other than English. As a reference, we have included the best performing unsupervised system for both Spanish and Italian (we will refer to this baseline as BestUns). This unsupervised baseline is based on the distributional models described in \citet{shwartz2017hypernymy}.

\begin{table}[!ht]
\resizebox{\textwidth}{!}{
\begin{tabular}{|l|l|rrr|rrr|}
\hline
\multirow{2}{*}{\textbf{Train data}} & \multicolumn{1}{c|}{\multirow{2}{*}{\textbf{Model}}} & \multicolumn{3}{c|}{\textbf{Spanish}}                                                    & \multicolumn{3}{c|}{\textbf{Italian}}                                                  \\ \cline{3-8} 
                            & \multicolumn{1}{l|}{}                       & \multicolumn{1}{l|}{\textit{MRR}} & \multicolumn{1}{l|}{\textit{MAP}} & \multicolumn{1}{l|}{$P@5$} & \multicolumn{1}{l|}{\textit{MRR}} & \multicolumn{1}{l|}{\textit{MAP}} & \multicolumn{1}{l|}{$P@5$} \\ \hline
-                           & BestUns                                     &    2.4                      &          5.5                &             2.5             &                    3.9      &           8.7               &        3.9                 \\ \hline \hline
\multirow{6}{*}{EN}         & VecMap\textsubscript{uns}                                   & 13.58 & 4.89 & 4.52 & 10.95 & 4.47 & 4.23                     \\
\cline{2-8}
                            & VecMap\textsubscript{multistep}                                   & 2.49 & 0.82 & 0.06 & 3.65 & 1.37 & 1.22                     \\
\cline{2-8}
                            &

                            VecMap                                  & 11.05 & 4.36 & 4.24 & 8.53 & 3.40 & 3.12                     \\
                            & Meemi (VecMap)                           & 14.62 & 5.56 & 5.43 & 11.50 & 4.52 & 4.36                     \\
                            & Meemi\textsubscript{w} (VecMap)                 & \textbf{15.33} & \textbf{5.89} & \textbf{5.69} & \textbf{13.11} & \textbf{5.08} & \textbf{4.72}                     \\
                            & Meemi-multi (VecMap)                     & 14.39 & 5.50 & 5.22 & 11.46 & 4.58 & 4.44                     \\ \hline
                            \hline
\multirow{6}{*}{EN + 500}   & VecMap\textsubscript{uns}                                   & 14.91 & 6.06 & 5.82 & 12.22 & 5.28 & 5.20                     \\
\cline{2-8}
                            &
                            
                            VecMap\textsubscript{multistep}                                   & 11.00 & 4.37 & 4.43 & 9.36 & 3.99 & 3.82                     \\
\cline{2-8}
                            &
                            
                            VecMap                                  & 12.20 & 5.00 & 4.93 & 9.95 & 4.17 & 4.08                     \\
                            & Meemi (VecMap)                           & 15.64 & 6.13 & 5.87 & 11.29 & 4.78 & 4.57                     \\
                            & Meemi\textsubscript{w} (VecMap)                 & \textbf{16.29} & \textbf{6.58} & \textbf{6.40} & \textbf{13.94} & \textbf{5.33} & \textbf{4.87}                     \\
                            & Meemi-multi (VecMap)                     & 15.03 & 6.20 & 6.26 & 12.46 & 4.88 & 4.60                     \\ \hline
                            \hline
\multirow{6}{*}{EN + 1K}      & VecMap\textsubscript{uns}                                   & 16.85 & 6.76 & 6.48 & 13.43 & 5.47 & 5.21                     \\
\cline{2-8}
                            & 
                            
                            VecMap\textsubscript{multistep}                                   & 12.39 & 4.95 & 4.88 & 11.95 & 5.22 & 5.03                     \\
\cline{2-8}
                            &
                            
                            VecMap                                  & 12.99 & 5.44 & 5.21 & 12.71 & 5.23 & 5.01                     \\
                            & Meemi (VecMap)                           & 17.46 & 6.82 & 6.43 & \textbf{14.53} & \textbf{5.92} & \textbf{5.74}                     \\
                            & Meemi\textsubscript{w} (VecMap)                 & \textbf{17.58} & \textbf{6.85} & 6.54 & 14.05 & 5.57 & 5.29                     \\
                            & Meemi-multi (VecMap)                     & 15.36 & 6.59 & \textbf{6.69} & 13.50 & 5.45 & 5.16                     \\ \hline
                            \hline
\multirow{6}{*}{EN + 2K}    & VecMap\textsubscript{uns}                                   & 16.44 & 6.83 & 6.53 & 14.04 & 6.03 & 5.90                     \\
\cline{2-8}
                            & 
                            
                            VecMap\textsubscript{multistep}                                   & 14.42 & 5.75 & 5.54 & 13.97 & 5.86 & 5.68                     \\
\cline{2-8}
                            &
                            
                            VecMap                                  & 14.59 & 6.24 & 6.21 & 13.10 & 5.63 & 5.40                     \\
                            & Meemi (VecMap)                           & \textbf{18.63} & \textbf{7.67} & \textbf{7.48} & \textbf{15.4} & \textbf{6.29} & \textbf{5.95}                     \\
                            & Meemi\textsubscript{w} (VecMap)                 & 17.52 & 6.96 & 6.76 & 14.4 & 5.86 & 5.60                     \\
                            & Meemi-multi (VecMap)                     & 17.17 & 6.90 & 6.78 & 14.29 & 5.83 & 5.45                     \\ \hline
\end{tabular}
}
\caption{Cross-lingual hypernym discovery results. In this case, VecMap = VecMap\textsubscript{ortho}.}
\label{tab:hyp}
\end{table}

As such, we report experiments (Table \ref{tab:hyp}) with training data only from English (11,779 hyponym-hypernym pairs), and \textit{enriched} models informed with relatively few training pairs (500, 1K, and 2K) from the target languages. Evaluation is conducted with the same metrics as in the original SemEval task, i.e.,  Mean Reciprocal Rank (MRR), Mean Average Precision (MAP), and precision at 5 ($P@5$). Specifically, MRR rewards the position of the first correct retrieved hypernym:

$$
\mbox{{MRR}} = \frac{1}{|Q|}\sum_{i=1}^{|Q|}\frac{1}{rank_i}
$$

\noindent where $Q$ is a sample of experiment runs and $rank_i$ refers to the rank position of the \textit{first} relevant outcome for the \textit{i}th run. However, in this hypernym discovery dataset, the vast majority of terms accept more than one correct hypernym, which is why MAP was considered as the official task metric in the SemEval task. This metric is defined as follows:

$$
\mbox{MAP} = \frac{1}{|Q|}\sum_{q \in Q} AP(q)
$$

\noindent where $AP$ (Average Precision) is the average of the $P@{K_1},...,P@{K_n}$ scores, where $K_1,...,K_n$ are the positions where the gold hypernyms appear in the ranking. As the maximum number of hypernyms allowed per term was 15, we only consider the first 15 gold hypernyms in cases where there are more.

We report comparative results between the following systems: VecMap\textsubscript{uns} (the unsupervised variant), VecMap\textsubscript{ortho} (the orthogonal transformation variant), VecMap\textsubscript{multi-step} (the supervised multi-stage variant) and three Meemi variants: Meemi (VecMap); Meemi\textsubscript{w} (VecMap) and Meemi-multi (VecMap). The first noticeable trend is the better performance of the unsupervised VecMap version versus its supervised orthogonal and multi-step counterparts. Nevertheless, we find remarkably consistent gains over both VecMap variants when applying Meemi, across all configurations for the two language pairs considered. In fact, the weighted (Meemi\textsubscript{w}) version brings an increase in performance between 1 and 2 MRR and MAP points across the whole range of target language supervision (from zero to 2k pairs). This is in contrast to the instrinsic evaluation, where the weighted model did not seem to provide noticeable improvements over the \textit{plain} version of Meemi. Finally, concerning the fully multilingual model, the experimental results suggest that, while still better than the orthogonal baselines, it falls short when compared to the weighted bilingual version of Meemi. This result suggests that exploring weighting schemes for the multilingual setting may bring further gains, but we leave this extension for future work.

\subsection{Cross-lingual natural language inference}
\label{XLI}

The task of natural language inference (NLI) consists in detecting entailment, contradiction or neutral relations in pairs of sentences. In our case, we test a zero-shot cross-lingual transfer setting where a system is trained with English corpora and is then evaluated on a different language. We base our approach on the assumption that better aligned cross-lingual embeddings should lead to better NLI models, and that the impact of the input embeddings may become more apparent in simple methods; as opposed to, for instance, complex neural network architectures. Hence, and also to account for the coarser linguistic granularity of this task (being a sentence classification problem rather than word-level), we employ a simple bag-of-words approach where a sentence embedding is obtained through word vector averaging. 
We then train a linear classifier\footnote{The codebase for these experiments is that of SentEval~\citep{conneau2018senteval}.} to predict one of the three possible labels in this task, namely \textit{entailment}, \textit{contradiction} or \textit{neutral}. We use the full MultiNLI English corpus \citep{williams2018broad} for training and the Spanish and German test sets from XNLI~\citep{conneau2018xnli} for testing. For comparison, we also include a lower bound obtained by considering English monolingual embeddings for input; in this case FastText trained on the UMBC corpus, which is the same model used to obtain multilingual embeddings.

Accuracy results are shown in Table~\ref{tab:xli}. The main conclusion in light of these results is the remarkable performance of the unsupervised VecMap model and, most notably, multilingual Meemi for both Spanish and German, clearly outperforming the orthogonal bilingual mapping baseline. Our results are encouraging for two reasons. First, they suggest that, at least for this task, collapsing several languages into a unified vector space is better than performing pairwise alignments. And second, the inherent benefit of having one single model accounting for an arbitrary number of languages.

\begin{table}[!ht]
\begin{center}
\resizebox{.55\textwidth}{!}{
\begin{tabular}{|l|r|r|}
\hline
\multicolumn{1}{|c|}{\textbf{Model}} & \multicolumn{1}{c|}{\textbf{EN-ES}} & \multicolumn{1}{c|}{\textbf{EN-DE}} \\ \hline
VecMap\textsubscript{uns} & 45.5 & 44.4 \\
\hline
VecMap\textsubscript{multistep} & 44.4 & 37.7  \\
\hline
VecMap\textsubscript{ortho}
& 43.9 & 43.6  \\
Meemi (VecMap\textsubscript{ortho}) & 44.9 & 43.8  \\
Meemi\textsubscript{w} (VecMap\textsubscript{ortho}) & 40.4 & 43.5 \\
Meemi-multi (VecMap\textsubscript{ortho}) & \textbf{46.6} & \textbf{45.5} \\
\hline
\hline
Lower bound & 38.0 & 33.4  \\ \hline
\end{tabular}
}
\end{center}
\caption{Accuracy on the XNLI task using different cross-lingual embeddings as features.}
\label{tab:xli}
\end{table}

\section{Analysis}

We complement our quantitative (intrinsic and extrinsic) evaluations with an 
analysis aimed at discovering the most salient characteristics of the transformation that is found by Meemi. 
We present a qualitative analysis with examples in Section \ref{qualitative}, as well as an analysis on the impact of the size of training dictionaries in Section \ref{influece_training} and on the performance of the multilingual model in Section \ref{multi-eval}.

\subsection{Studying word translations}
\label{qualitative}

Table \ref{tab:examples} lists a number of examples where, for a source English word, we explore its highest ranked \textit{cross-lingual synonyms} (or word translations) in a target language. We select Spanish as a use case. 

\begin{table}[!ht]
\resizebox{\textwidth}{!}{
\begin{tabular}{|ccc|ccc|}
\hline 
\multicolumn{3}{|c|}{\textbf{crazy}}                                                                                                    & \multicolumn{3}{c|}{\textbf{telegraph}}                                                                                               \\ \hline
\multicolumn{1}{|c|}{\textbf{VecMap}} & \multicolumn{1}{c|}{\textbf{Meemi}} & \multicolumn{1}{|c|}{\textbf{Meemi-multi}} & \multicolumn{1}{c|}{\textbf{VecMap}} & \textbf{Meemi}                      & \textbf{Meemi-multi}                      \\ \hline
\textbf{loco}                                 & \textbf{loco}                                        & chifladas                                        & \textbf{tel\'{e}grafo}                            & telegr\'{a}fico                                 & \textbf{telegraph}                                        \\
tonto                                & \textbf{loca}                                        & \textbf{locos}                                            & tel\'{e}grafos                           & \textbf{tel\'{e}grafo}                                   & telegraaf                                        \\
enloquecere                          & enloquec\'{i}                                   & \textbf{loca}                                             & telegr\'{a}fico                          & telegr\'{a}fono                                 & telegraphone                                     \\
\textbf{locos}                                & enloquec\'{i}as                                 & est\'{u}pidas                                        & telegr\'{a}fica                          & telegraf                                    & telegr\'{a}fono                                      \\
enloqueci                            & \textbf{locos}                                       & alocadas                                         & telegrafo                            & telegr\'{a}fo                                   & \textbf{tel\'{e}grafo}                                        \\ \hline \hline
\multicolumn{3}{|c}{\textbf{conventions}}                                                                                              & \multicolumn{3}{|c|}{\textbf{discover}}                                                                                                \\ \hline
\multicolumn{1}{|c|}{\textbf{VecMap}} & \multicolumn{1}{c|}{\textbf{Meemi}} & \multicolumn{1}{c|}{\textbf{Meemi-multi}} & \multicolumn{1}{c|}{\textbf{VecMap}} & \multicolumn{1}{c|}{\textbf{Meemi}} & \textbf{Meemi-multi}                      \\ \hline
\textbf{convenciones}                         & internaciones                               & \textbf{convenios}                                        & descubrir\'{a}                           & \textbf{descubre}                                    & descubr                                          \\
internacional7                       & 1972naciones                                & reglas                                           & descubr                              & \textbf{descubrir}                                   & descubrir\'{a}n                                      \\
convenci\'{o}n                           & protocolos                                  & convenci\'{o}n                                       & descubrir\'{a}n                          & descubriendo                                & descubrirnos                                     \\
1961naciones                         & \textbf{convenios}                                   & normas                                           & \textbf{descubren}                            & \textbf{descubra}                                    & descubrira                                       \\
internacionales3                     & 1961naciones                                & legislacionesnacionales                          & descubriron                          & descubrira                                  & descubrire                                       \\ \hline \hline
\multicolumn{3}{|c|}{\textbf{remarks}}                                                                                                  & \multicolumn{3}{c|}{\textbf{lyon}}                                                                                                    \\ \hline
\multicolumn{1}{|c|}{\textbf{VecMap}} & \multicolumn{1}{c|}{\textbf{Meemi}} & \multicolumn{1}{c|}{\textbf{Meemi-multi}} & \multicolumn{1}{c|}{\textbf{VecMap}} & \multicolumn{1}{c|}{\textbf{Meemi}} & \multicolumn{1}{c|}{\textbf{Meemi-multi}} \\ \hline
astrom\'{e}tricos                        & lobservaciones                              & \textbf{observaciones}                                    & rocquigny                            & beaubois                                    & marcigny                                         \\
observacionales                      & mediciones                                  & observacionales                                  & r\'{e}milly                              & bourgmont                                   & \textbf{lyon}                                             \\
astrom\'{e}tricas                        & lasobservaciones                            & observacional                                    & martignac                            & marcigny                                    & pierreville                                      \\
astronom\'{e}tricas                      & deobservaciones                             & predicciones                                     & beaubois                             & r\'{e}milly                                     & jacquemont                                       \\
predicciones                         & susobservaciones                            & mediciones                                       & chambourcy                           & jacquemont                                  & beaubois                      \\ \hline                  
\end{tabular}
}
\caption{Word translation examples from English and Spanish, comparing VecMap with the bilingual and multilingual variants of Meemi. For each source word, we show its five nearest cross-lingual synonyms. Bold translations are correct, according to the source test dictionary (cf. Section \ref{induction}).}
\label{tab:examples}
\end{table}

Let us study the examples listed in Table \ref{tab:examples}, as they constitute illustrative cases of linguistic phenomena which go beyond correct or incorrect translations. First, the word 'crazy' is correctly translated by both VecMap and Meemi; \textit{loco} (masculine singular), \textit{locos} (masculine plural) or \textit{loca} (feminine) being standard translations, with no further connotations, of the source word. However, the most interesting finding lies in the fact that for Meemi-multi, the preferred translation is a colloquial (or even vulgar) translation which was not considered as correct in the gold test dictionary. The Spanish word \textit{chifladas} translates to English as `going mental' or `losing it'. 
Similarly, we would like to highlight the case of `telegraph'. This word is used in two major senses, namely to refer to a message transmitter and as a reference to media outlets (several newspapers have the word `telegraph' in their name). VecMap and Meemi (correctly) translate this word into the common translation \textit{tel\'{e}grafo} (the transmission device), whereas Meemi-multi prefers its named-entity sense. 

Other cases, such as `conventions' and `discover' are examples to illustrate the behaviour for common ambiguous nouns. In both cases, candidate translations are either misspellings of the correct translation (\textit{descubr} for `discover'), or misspellings involving tokens conflating two words whose compositional meaning is actually a correct candidate translation for the source word; e.g., \textit{legislaciones nacionales} (`national rulings') for `conventions'. Finally, `remarks' offers an example of a case where ambiguity causes major disruptions. In particular, `remark' translates in Spanish to \textit{observaci\'{o}n}, which in turn has an astronomical sense; `astronomical observatory' translates to \textit{observatorio astron\'{o}mico}. 

\subsection{Impact of training dictionary and corpus size}
\label{influece_training}

\begin{figure}[!h]
    \centering
    \subfloat[Meemi (VecMap)]{{\includegraphics[width=11cm]{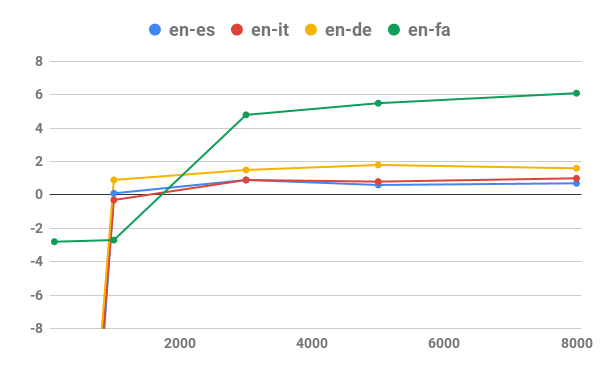} }}%
    \qquad
    \subfloat[Meemi (MUSE)]{{\includegraphics[width=11cm]{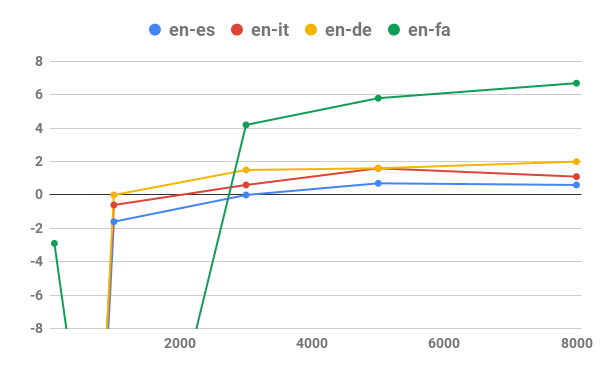} }}%
    \caption{Absolute improvement (in terms of Pearson correlation percentage points) by applying the Meemi over the two base orthogonal models VecMap and MUSE on the cross-lingual word similarity task, with different training dictionary sizes. As data points in the X-axis we selected 100, 1000, 3000, 5000 and 8000 word pairs in the dictionary.}
    \label{fig:dictsize}%
\end{figure}

Our method relies on the availability of suitable bilingual training dictionaries, where we can expect that the size of these dictionaries should have a clear impact on the quality of the final transformation. This is analyzed in Figure \ref{fig:dictsize} for the task of cross-lingual word similarity. The figure shows the absolute improvement (in percentage points) over VecMap by applying Meemi, using different training dictionary sizes for supervision.

As can be observed, using Meemi improves the results, for all language pairs, when dictionaries of 8K, 5K or 3K word pairs are used, but its performance heavily drops with dictionaries of smaller sizes (i.e.\ 1K and especially 100). In fact, having a larger dictionary helps avoid overfitting, which is a recurring problem in cross-lingual word embedding learning \citep{zhang2017adversarial}. The most remarkable case is that of Farsi, where Meemi improves the most, but where access to a sufficiently large dictionary becomes even more important. This behavior clearly shows under which conditions our proposed final transformation can be applied with higher success rates. We leave exploring larger dictionaries and their impact in different tasks and languages for future work.

On the other hand, we have observed that while corpus size plays a role in the performance of our models, it is not as notable as it might seem at first.
Given the different corpus sizes of the data we used to train our monolingual embeddings, we analyzed the correlation between these sizes, mentioned in Section~\ref{corpora}, and the performance figures presented in Tables~\ref{tab:dictinduction}, \ref{tab:cross-sim}, and~\ref{tab:mono-sim}.
The average Pearson correlation across multilingual models in dictionary induction, where all languages are available, is 0.38 (discarding VecMap\textsubscript{uns} due to its anomalous results for Finnish), while for cross-lingual and monolingual word similarity it is 0.69 and 0.65, respectively. Note, however, that in these latter cases we are missing two distant languages, i.e., Finnish and Russian.

\subsection{Multilingual performance}
\label{multi-eval}

In this section we assess the benefits of our proposed multilingual integration (cf. Section \ref{multilingual}). 
To this end, we measure fluctuations in performance as more languages were added to the initially bilingual model. 
Thus, starting from a bilingual embedding space obtained with VecMap\textsubscript{ortho}, we apply Meemi over a number of aligned spaces, which ultimately leads to a fully multilingual space containing the following languages: Spanish, Italian, German, Finnish, Farsi, Russian, and English. This latter language is used as the target embedding space for the orthogonal transformations due to it being the richest in terms of resource availability.

To avoid a lengthy and overly exhaustive approach where all possible combinations from two to seven languages are evaluated, we opted for conducting an experiment where languages are divided into two groups and added one by one in a fixed order: the first group is formed by languages that obtain the best alignments with English in previous experiments, which broadly coincides with those that are closer to English in terms of language family and alphabet (i.e., Spanish, Italian, and German), and then the second group formed by the remaining languages (i.e., Finnish, Farsi, and Russian).
However, this approach does not allow us to use, for example, the English-Farsi test set until reaching the fifth step.
To solve this, if the language that is needed for the test set has not yet been included, we replace the last language that was added by the one that is needed for the test set.
For instance, while we normally add Italian as the second source language (resulting in trilingual space \textit{en}-\textit{es}-\textit{it}), for the English-German test set, the results are instead based on a space where we added German instead of English (i.e.\ the trilingual space \textit{en}-\textit{es}-\textit{de}).
In Table~\ref{tab:multilingual} we show the results obtained by the multilingual models in bilingual dictionary induction. 

\begin{table}[!t]
\resizebox{\textwidth}{!}{
\begin{tabular}{|l|rrr|rrr|rrr|}
\hline
\multirow{2}{*}{\textbf{Languages}} & \multicolumn{3}{c|}{\textbf{English-Spanish}} & \multicolumn{3}{c|}{\textbf{English-Italian}} & \multicolumn{3}{c|}{\textbf{English-German}} \\ \cline{2-10} 
 & \multicolumn{1}{c|}{\textbf{$P@1$}} & \multicolumn{1}{c|}{\textbf{$P@5$}} & \multicolumn{1}{c|}{\textbf{$P@10$}} & \multicolumn{1}{c|}{\textbf{$P@1$}} & \multicolumn{1}{c|}{\textbf{$P@5$}} & \multicolumn{1}{c|}{\textbf{$P@10$}} & \multicolumn{1}{c|}{\textbf{$P@1$}} & \multicolumn{1}{c|}{\textbf{$P@5$}} & \multicolumn{1}{c|}{\textbf{$P@10$}} \\ \hline
$x$-en (VecMap\textsubscript{ortho}) & 32.6 & 58.1 & 65.8 & 32.9 & 56.5 & 63.4 & 22.8 & 42.8 & 50.4 \\ \hline
$x$-en & 33.9 & 60.7 & 67.4 & 33.8 & 58.8 & 65.6 & 23.7 & 45.0 & 52.9 \\
es-$x$-en & \textbf{34.2} & 60.8 & 68.2 & 33.3 & 58.1 & 66.5 & \textbf{23.9} & \textbf{45.9} & 53.2 \\
es-it-$x$-en & 34.1 & 61.2 & 68.1 & 33.8 & \textbf{58.9} & \textbf{66.7} & 23.8 & 45.8 & 53.1 \\
es-it-de-$x$-en & \textbf{34.2} & \textbf{61.3} & \textbf{68.3} & \textbf{33.9} & 58.8 & 66.5 & \textbf{23.9} & 45.6 & \textbf{53.4} \\
es-it-de-fi-$x$-en & 33.6 & 60.9 & 67.5 & 33.8 & 58.0 & 65.8 & 23.1 & 44.7 & 52.7 \\
es-it-de-fi-fa-ru-en & 33.4 & 60.9 & 67.1 & 33.7 & 58.1 & 65.5 & 23.0 & 44.5 & 52.8 \\ \hline \hline
 & \multicolumn{3}{c|}{\textbf{English-Finnish}} & \multicolumn{3}{c|}{\textbf{English-Farsi}} & \multicolumn{3}{c|}{\textbf{English-Russian}} \\ \cline{2-10} 
 & \multicolumn{1}{c|}{\textbf{$P@1$}} & \multicolumn{1}{c|}{\textbf{$P@5$}} & \multicolumn{1}{c|}{\textbf{$P@10$}} & \multicolumn{1}{c|}{\textbf{$P@1$}} & \multicolumn{1}{c|}{\textbf{$P@5$}} & \multicolumn{1}{c|}{\textbf{$P@10$}} & \multicolumn{1}{c|}{\textbf{$P@1$}} & \multicolumn{1}{c|}{\textbf{$P@5$}} & \multicolumn{1}{c|}{\textbf{$P@10$}} \\ \hline
$x$-en (VecMap\textsubscript{ortho}) & 22.1 & 44.5 & 52.9 & 18.5 & 33.6 & 40.5 & 15.6 & 35.5 & 44.2 \\ \hline
$x$-en & 24.2 & 48.8 & 57.7 & 20.0 & 37.1 & 43.8 & 19.0 & 40.5 & 49.9 \\
es-$x$-en & \textbf{24.7} & 50.1 & 58.4 & 21.1 & \textbf{37.9} & 43.9 & 17.9 & 40.2 & 49.3 \\
es-it-$x$-en & 24.1 & \textbf{51.1} & \textbf{59.2} & 20.9 & 37.6 & 44.5 & 18.9 & 41.6 & 50.6 \\
es-it-de-$x$-en & 23.9 & 50.2 & 58.5 & 21.0 & 37.7 & \textbf{44.9} & 18.9 & 41.5 & 50.8 \\
es-it-de-fi-$x$-en & 23.5 & 48.6 & 57.5 & \textbf{21.2} & 37.5 & 44.0 & \textbf{19.1} & \textbf{42.1} & \textbf{51.4} \\
es-it-de-fi-fa-ru-en & 23.1 & 48.3 & 57.2 & 21.0 & \textbf{37.9} & 44.4 & 18.8 & 41.7 & 50.5 \\
\hline
\end{tabular}
}
\caption{Dictionary induction results obtained with the multilingual extension of Meemi over (VecMap\textsubscript{ortho}). The sequence in which source languages are added to the multilingual models is: Spanish, Italian, German, Finnish, Farsi, and Russian (English is the target). The $x$ indicates the use of the test language in each case (if the test language is already included, the following language in the sequence is added). 
We also include the scores of the original VecMap\textsubscript{ortho} as baseline.}
\label{tab:multilingual}
\end{table}

The best results are achieved when more than two languages are involved in the training, which correlates with the results obtained in the rest of the tasks and highlights the ability of Meemi to successfully exploit multilingual information to improve the quality of the embedding models involved. In general, the performance fluctuates more significantly when adding the first language to the bilingual models and then stabilizes at a similar level to the bilingual case when adding more distant languages. 

\section{Conclusion}

In this article, we have presented an extended study of Meemi, a simple post-processing method for improving cross-lingual word embeddings which was first presented in \citet{doval:meemiemnlp2018}. 
Our initial goal was to learn improved bilingual alignments from those obtained by state-of-the-art cross-lingual methods such as VecMap \citep{artetxe2018generalizing} or MUSE \citep{conneau2018word}. We do this by applying a final unconstrained linear transformation to their initial mappings. Our extensive evaluation reveals that Meemi, using only dictionary translation as supervision, can improve on the supervised and unsupervised variants of these models, in both close and distant languages. This also confirms findings from recent work that unsupervised models may be more brittle than supervised models, even if these are using only word translations as supervision \citep{vulic2019we}. 

In this work, we have also gone beyond the bilingual setting by exploring an extension of the original Meemi model to align embeddings from an arbitrary number of languages in a single shared vector space. 
In particular, we take advantage of the fact that, assuming the initial alignment was obtained with an orthogonal mapping, Meemi can naturally be applied to any number of languages through a single linear transformation per language. 

Regarding the evaluation, we extended the language set to include, in addition to the usual Indo-European languages such as English, Spanish, Italian or German, other distant languages such as Finnish, Farsi, and Russian. 
The results we report in this article show that Meemi is highly competitive, consistently yielding better results than competing baselines, especially in the case of distant languages. We are particularly encouraged by the multilingual results, which prove that bringing together distant languages from different families in a shared vector space appears to be beneficial in most cases.  

\section{Future Work}

We will continue to explore the possibilities of post-processing multilingual models, investigating their impact in different tasks.  
Given the fact that going from restrictive orthogonal transformations to the less constrained Meemi transformation was found to be beneficial in the integration of monolingual models, it remains to be seen whether there are benefits in further fine-tuning the alignment, in the form of some kind of constrained non-linear transformation. 

Given the recent breakthroughs in multilingual contextualized language models such as mBERT~\citep{devlin2019bert}, we also plan on exploring the use of static (i.e.,\ non-contextualized) cross-lingual word embeddings as prior knowledge for those models, as was suggested by \citet{artetxe20transferability} (see ending of Section~\ref{intro}). 
More specifically, instead of freezing the pretrained input embeddings when training the contextualized model, it would be interesting to analyze the effect of updating the parameters of the cross-lingual word vectors jointly with the rest of the language model.
An advantage of our cross-lingual vectors, compared to the ones that were considered by \citet{artetxe20transferability}, is that we can train them on a wider range of languages (i.e., not just bilingual), which would allow for a more comprehensive exploitation of multilingual training corpora. 

\section*{Acknowledgments}
Yerai Doval has been supported by the Spanish Ministry of Economy, Industry and Competitiveness (MINECO) through the ANSWER-ASAP project (TIN2017-85160-C2-2-R); by the Spanish State Secretariat for Research, Development and Innovation (which belongs to MINECO) and the European Social Fund (ESF) through a FPI fellowship (BES-2015-073768) associated to TELEPARES project (FFI2014-51978-C2-1-R); and by the Xunta de Galicia through TELGALICIA research network (ED431D 2017/12).
This work was partly supported by ERC Starting Grant 637277.

\bibliography{mybibfile}
\bibliographystyle{nlelike}

\end{document}